\documentclass[12pt, a4paper]{article}

\usepackage[a4paper,margin=20mm]{geometry}
\usepackage[utf8]{inputenc}
\usepackage{amsmath}
\usepackage{amssymb}

\usepackage{graphicx}
\usepackage{color}

\begin{document}


\newcommand{\HRule}{\rule{\linewidth}{0.5mm}} 
\begin{center}
\HRule \\[0.4cm]
{\huge\bfseries Artificial SA-I, RA-I and RA-II/Vibrotactile Afferents for Tactile Sensing of Texture}\\[0.4cm] 
\HRule \\[1.5cm]
\textbf{Nicholas~Pestell and~Nathan~F.~Lepora}\\[0.5cm]
Department of Engineering Mathematics and Bristol Robotics Laboratory,\\ University of Bristol, BS8 1QU, UK\\
e-mail: n.lepora@bristol.ac.uk\\[0.5cm]
{\large \today}\\[3cm]

\begin{abstract}
Robot touch can benefit from how humans perceive tactile textural information, from the stimulation mode to which tactile channels respond, then the tactile cues and encoding. Using a soft biomimetic tactile sensor (the TacTip) based on the physiology of the dermal-epidermal boundary, we construct two biomimetic tactile channels based on slowly-adapting SA-I and rapidly-adapting RA-I afferents, and introduce an additional sub-modality for vibrotactile information with an embedded microphone interpreted as an artificial RA-II channel. These artificial tactile channels are stimulated dynamically with a set of 13 artificial rigid textures comprising raised-bump patterns on a rotating drum that vary systematically in roughness. Methods employing spatial, spatio-temporal and temporal codes are assessed for texture classification insensitive to stimulation speed. We find: (i) spatially-encoded frictional cues provide a salient representation of texture; (ii) a simple transformation of spatial tactile features to model natural afferent responses improves the temporal coding; and (iii) the harmonic structure of induced vibrations provides a pertinent code for speed-invariant texture classification. Just as human touch relies on an interplay between slowly-adapting (SA-I), rapidly-adapting (RA-I) and vibrotactile (RA-II) channels, this tripartite structure may be needed for future robot applications with human-like dexterity, from prosthetics to materials testing, handling and manipulation.       
\end{abstract}

\newpage
\end{center}

\section{Introduction}
\label{sec01}
Human tactile texture sensation is likely governed by complex relationships between the physical dimensions and properties of the texture and the nature of the mode of stimulation. Researchers have proposed a range of mechanisms for encoding properties such as roughness \cite{sathian,cascio} in peripheral neural activity, encompassing different tactile channels (afferent types) and representations within spike trains (rate codes \cite{goodwin2}, population codes \cite{connor}, and temporal codes \cite{connor}). \textcolor{black}{That said}, a combination of afferent types and encoding schemes are likely employed to form a complete tactile portrait of texture, as has been observed with other tactile dimensions \cite{saal}.

Tactile texture perception has been examined widely in robotics with primary focus on classification of discrete texture classes, \textcolor{black}{commonly with bioinspired sensors employing a vibrotactile modality \cite{mukaibo2005,sinapov2011,fishel,yi2017}}. The trend has been towards using complex natural textures \cite{li2,baishya,romano2} that vary across a range of dimensions, such as roughness, hardness and stickiness \cite{fishel}. The focus in robotics has been on achieving high-accuracy on natural textures, for example by using large data-sets and high-resolution data. \textcolor{black}{Using natural textures does have its drawbacks though. For example, it opens a disconnect with investigations of human touch where there is a long tradition of using artificial stimuli that vary systematically in a single perceptual dimension such as roughness.} 

This approach towards high-accuracy texture classification is exemplified in \cite{taunyazov}, where a range of techniques were examined for classifying 23 natural textures using the iCub's tactile forearm; the best performing method was a combined CNN-LSTM neural network model trained on spatio-temporal pressing and sliding data. Another example of texture classification is with the GelSight optical tactile sensor, where fine-grained spatial information is encoded in high-resolution tactile images~\cite{li2}. Whilst these methods are clearly capable of achieving high accuracy on raw data from the sensors, \textcolor{black}{they were not intended to relate to how humans perceive texture or be data efficient}. 

These robotic approaches are also orthogonal to investigations of the fundamental aspects of texture perception being considered in biology~\textcolor{black}{(e.g.~\cite{cascio2001,johnson2002,bensmaia2003})}. For artificial tactile sensing, as in nature, one might expect that the ability of systems to transduce textural information depends on a complex relationship between the stimulus scale, the mode of stimulation, the properties of the tactile channel used to transduce information, and the dimensionality of the tactile data. \textcolor{black}{Thus, to achieve artificial texture perception with human-like capabilities, it may be necessary to instantiate artificial tactile sensing that mimics the various biological modalities that comprise natural touch sensing}.

\begin{figure}[h!]
	\begin{center}
        \includegraphics[width=\textwidth, trim=0cm .5cm 0cm 0cm]{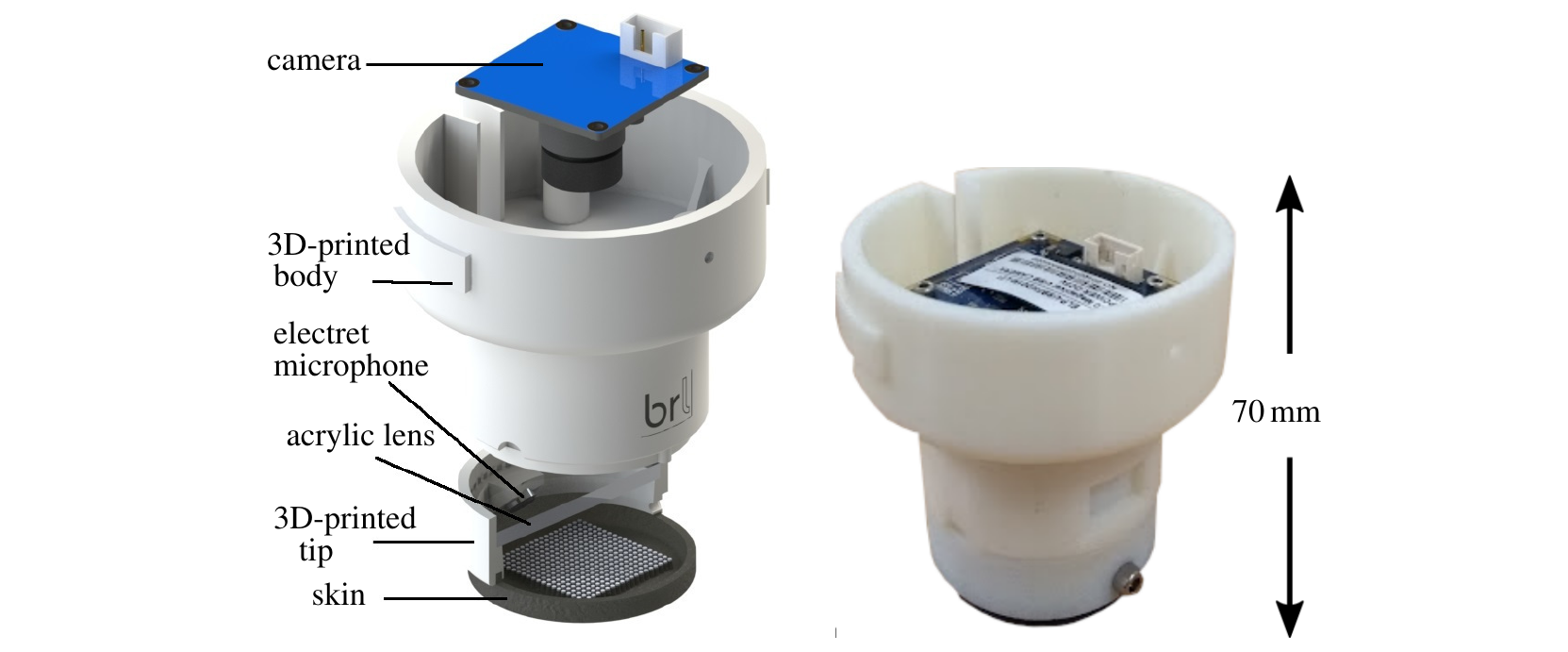}
	\end{center}
	\vspace{-.5em}
	\caption{Left panel: Exploded view of the TacTip soft biomimetic optical tactile sensor with integrated vibration sense \textcolor{black}{using the electret microphone}. Right panel: Photograph of a complete TacTip showing the physical scale. \textcolor{black}{The compliant skin of the sensor tip is a 40\,mm-diameter flat pad, with an array of markers on the inner surface visible to the camera. \textcolor{black}{These markers are illuminated with LEDs on a ring-shaped circuit board that fits between the 3D-printed tip and body.}}}
	\vspace{0em}
	\label{FP}
	\begin{center}
        \includegraphics[width=\textwidth, trim=0cm 1cm 0cm 0cm]{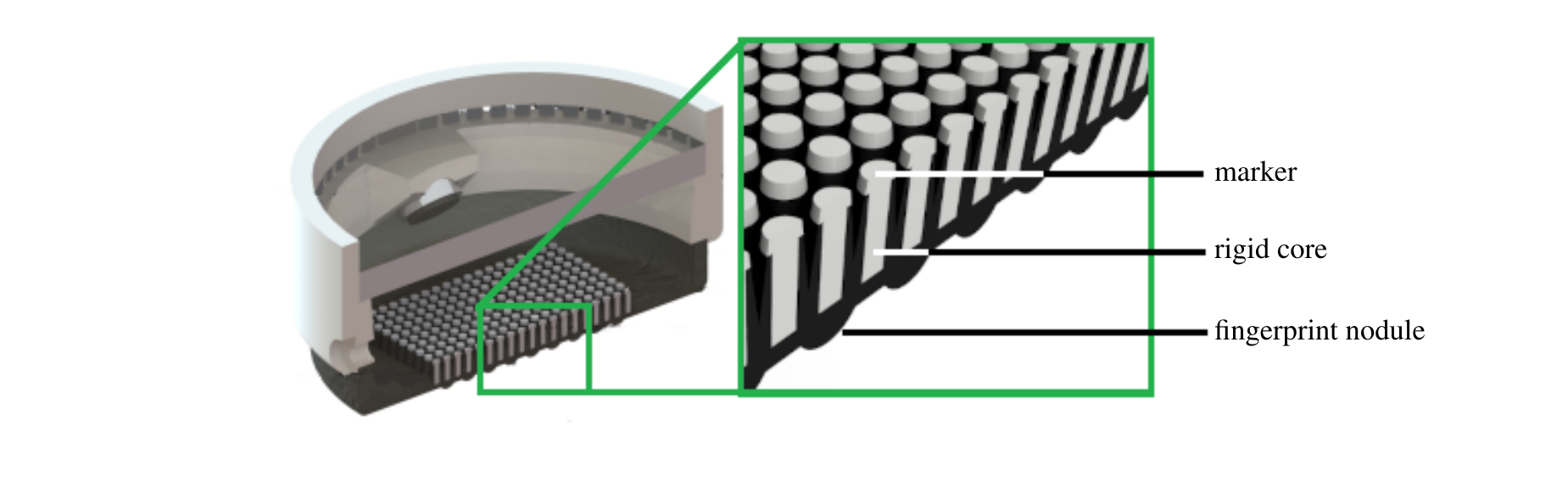}
	\end{center}
	\vspace{-.5em}
	\caption{\textcolor{black}{Cross section through the 3D-printed tip of the soft biomimetic optical tactile sensor shown in Figure 1. The arrangement of fingerprint nodules, markers and rigid cores of each movable pin are shown in the highlighted region. Fingerprint nodules are arranged along the diagonals of the pin array, so that \textcolor{black}{one} in every \textcolor{black}{two} pins has a nodule.}}
	\vspace{0em}
	\label{sensor_MM}
\end{figure}

To examine this proposal, we use an established soft biomimetic optical tactile fingertip, the TacTip (Figure \ref{sensor_MM}) \cite{Ward-Cherrier2018,lepora2021}, from which we extract three distinct feature sets proposed to model natural tactile channels in human touch; \textcolor{black}{these features are then used} to classify a set of 13 artificial textures via dynamic stimulation. Two of these feature sets are here referred to as artificial SA-I and RA-I afferents, because a related paper demonstrates \textcolor{black}{that they give viable models of the corresponding natural afferents for representing spatial information such as grating orientation \cite{pestell_gratings}. This study considers a fuller set of artificial SA-I, RA-I and RA-II tactile afferents}, with the main novel contributions: \\
\noindent(i) The introduction of a novel \textit{vibrotactile channel} to the TacTip, proposed to mimic natural RA-II tactile afferents, \textcolor{black}{by using a microphone embedded within the soft inner gel of the tactile sensor}.\\
\noindent(ii) A method for speed-insensitive texture perception based on using frictional cues encoded within spatial modulation of artificial afferent response, tested with textured stimuli comprising raised-dot patterns similar to those used to vary roughness in perceptual psychophysics experiments~\cite{connor}.\\
\noindent(iii) A method for speed-insensitive texture perception in the vibrotactile channel based on a theory of speed-invariant texture perception in humans, in that texture is signalled by the harmonic structure of induced vibrations encoded within the subsequent harmonic structure of afferent firing \cite{yau,saal2}.    

The results from our experiments identify new principles for improving robot touch, particularly in terms of data-efficient methods for \textcolor{black}{transducing tactile sensory information and extracting biologically-inspired tactile features} that could both transcend the particular type of tactile sensor and extend to tactile dimensions beyond texture classification.

\section{Methods}
\label{sec02}

\subsection{Sensor Design and Fabrication}
\label{sec02:01}

The relevant operating principles of the BRL TacTip are described below in relation to the customization of the design for texture sensing. For manufacturing and detailed explanations of the design concepts, we refer the reader to a 2018 paper on `The TacTip Family' \cite{Ward-Cherrier2018} and a recent (2021) review of `Soft Biomimetic Optical Tactile Sensing with the TacTip' \cite{lepora2021}.

The BRL TacTip has two main sections  \textcolor{black}{(Figures \ref{FP}, \ref{sensor_MM})}: a rigid body housing the electronic components, which is 3D-printed in ABS, and a soft tip fabricated using a multi-material FDM printer. The tip is comprised of two materials printed as a single part: a rubber-like `skin' (Tango Black Plus, Stratasys) and rigid rim (VeroWhite, Stratasys). The inside of the tip is \textcolor{black}{filled by injecting} a clear silicone gel (RTV27905, Techsil UK). The gel \textcolor{black}{provides} stiffness to the tip which helps to minimise hysteresis while enabling compliance with a stiffness difference between the skin (Shore A 26-28) and gel (Shore OO 10). \textcolor{black}{The gel is optically clear so that 3D-printed white markers on the inside of the compliant skin can be imaged by an internal camera.}


The TacTip body's primary function is to house electronic components and to serve as a mount for auxiliary robotic systems. To capture images of the white markers, a USB camera module (Ailipu Technology) is used \textcolor{black}{at resolution $640\times480$ pixels and frame rate 90\,fps}. The length of the TacTip body is chosen to provide full view of the markers through the wide-angle lens. At the base of the \textcolor{black}{body where it joins with the tip} is a PCB ring containing 6 white LEDs that illuminate the markers for imaging by the camera.

The sensor modifications for this work are as follows.

\indent {\bf Microphone Transduction:} In accordance with the aforementioned motivations, multi-modality is facilitated by leveraging acoustic vibrations occurring within the gel during dynamic stimulation. To detect these vibrations, a small electret microphone of diameter 6\,mm (Kingstate, KECG3642TF-A) is fitted to the inside of the TacTip's Vero White rim (see Figure \ref{sensor_MM}). During assembly, the microphone is press fitted into a cavity prior to filling with gel. The gel is in contact with the \textcolor{black}{front surface of the microphone} and the cabling is routed through a hole in the side of the tip. \textcolor{black}{The sensor is designed so that} physical interaction between the TacTip's `skin' and a stimulus will induce vibrations that propagate through the silicone gel, \textcolor{black}{which can then be picked up and} converted to a voltage by the microphone. The signal is amplified to line-level ($V$\textsubscript{pp} 2\,V) using an inverting op-amp circuit.
    
\indent {\bf Tip Shape:} In this work, a flat skin is used to facilitate contact with textured surfaces. The rationale behind this modification is to reduce the physical distance between the point of interface with stimuli and the microphone. Acoustic vibrations will attenuate as they travel through the silicone. In practice, after the tip has been filled with gel the skin is no longer completely flat, even though efforts were taken to maintain a flat profile during fabrication. In particular, gravity causes the skin surface to slightly `bulge’ due to the weight of the gel when the sensor is oriented downwards.
    
\indent {\bf Marker Layout:} The TacTip design here uses 361 pins/markers that is about three-times finer than the conventional TacTips used in other studies~\cite{Ward-Cherrier2018}. This design was both to improve spatial resolution and to have a common pin array with \cite{pestell_gratings}. The pins are arranged in a simple 19$\times$19 square array with separation 1.2\,mm (Figure \ref{FP}). A regular square array is used so that information encoded in the relationship between movements of neighbouring markers (spatial coding) can be more easily identified. The marker density of $\sim$70\,cm\textsuperscript{-2} is approximately half the innervation density of type-I afferents in the human fingertip \cite{johansson2,johansson3}.
    
\indent {\bf Artificial Fingerprint:} Scheibert et al.\ suggest that the human fingerprint may amplify the vibrations produced at the interface of the skin and stimulus, and so benefit texture perception \cite{scheibert}. Our preliminary investigations also confirmed that this effect occurs with our proposed artificial fingertip design. Here we leverage the design concepts realised by Cramphorn et al.\ (2017) \cite{Cramphorn2017} to similarly modify our multi-modal design to mimic the function of an external fingerprint and internal epidermal ridges. Specifically: \textit{(a)} The outside of the skin is augmented with small nodules (diameter 2\,mm) that form the physical analogue of the papillary ridges that comprise a fingerprint; these nodules are made from the same material as the skin and are thus also compliant. The pattern of nodules on the skin exterior mirrors that of the pins on the interior, with the nodules located directly below one-in-two pins (Figure~\ref{sensor_MM}). \textit{(b)} Each pin contains a rigid `core' that is mechanically fused with its respective white marker. The function of these cores is to enhance the stiffness contrast between the pins and the silicone gel to mimic the function of the stiffer epidermal ridges~\cite{gerling}. 


\begin{figure}[t]
	\centering
    \includegraphics[width=1\textwidth]{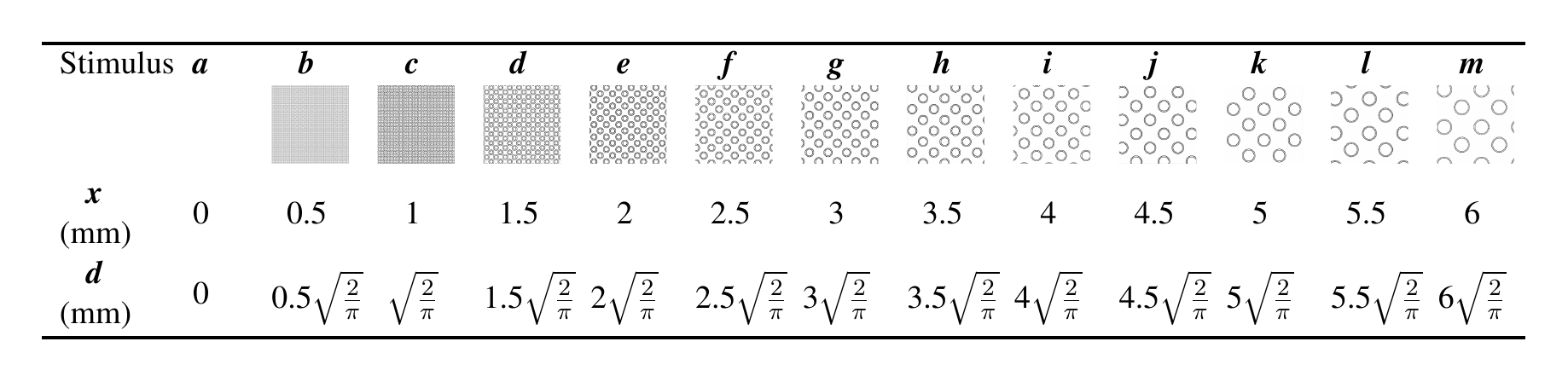}
	\vspace{-1em}
    \caption{Patterns of 13 textured stimuli, from completely smooth (\textbf{a}) to coarsest texture (\textbf{m}). \textcolor{black}{The textures comprise raised bumps of diameter $d$ and bump-to-bump spacing $x$ from one centre to another along the diagonals. The values $(x,d)$ are chosen to maintain the same overall bump area for all non-smooth patterns, so that contact area is not a stimulus cue.}}
	\vspace{-1em}
	\label{table1}
	\vspace{3em}
	\centering
    \includegraphics{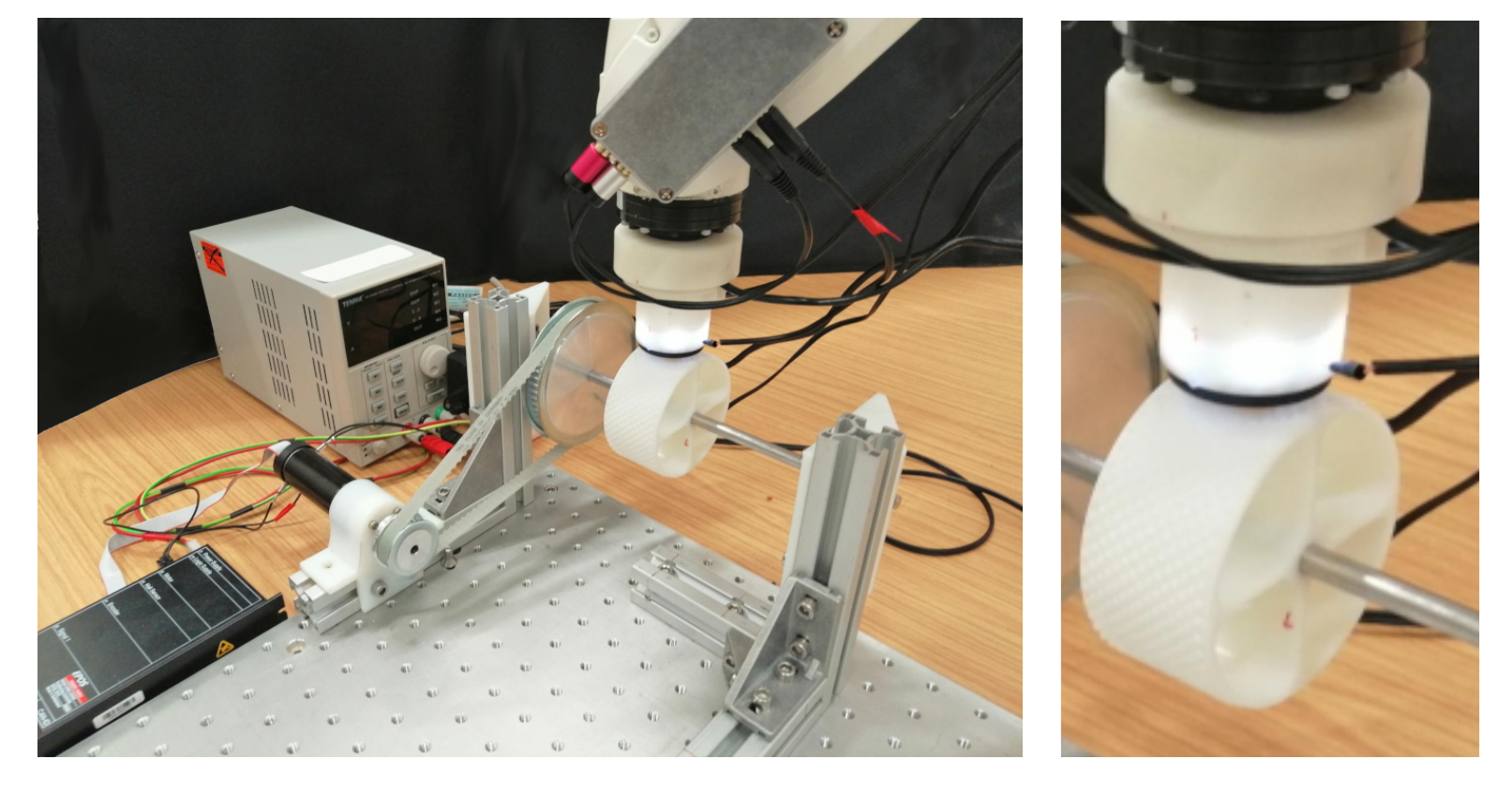}
    \vspace{-1em}
	\caption{Experimental set up for dynamic touch experiments with textured stimuli. The tactile sensor is held stationary with its skin against a rotating textured drum whose speed is modulated by a DC motor and drive mechanism.} 
	\label{texture_rig}
\end{figure}
\noindent

\subsection{Data Collection}
\label{sec02:03}

For experimental testing, a set of textured drum stimuli are used that are patterned with tetragonal arrays of raised bumps on their exterior circumferences (Figure~\ref{table1}, $N_{\rm{tex}}$ = 13). The drums are 3D-printed in a rigid plastic (VeroWhite, Stratasys) using a PolyJet printer (Objet, Stratasys), with the textured bumps raised 0.5\,mm beyond the drum's 80\,mm diameter. Both the diagonal bump separation, $x$, and the bump diameter, $d$, are varied systematically  to provide a linear variation in texture period whilst constraining the area of raised bumps across stimuli to be constant (Figure~\ref{table1}). This constraint is important because it prohibits the availability of intensive cues due to surface area that would be available from the contact, rather than the roughness of the texture.

The tactile sensor is held stationary against textured drums (Figure \ref{texture_rig}), which are driven by a 60\,W brushed motor connected to the drum spindle. This connection uses a 4.8:1 planetary gear head and a custom-built 3:1 belt drive. The drum velocity is controlled in closed loop.

During the experiments, the sensor is held with interpenetration depth close to 1\,mm and pointing vertically downwards, \textcolor{black}{with the apparatus designed so that this indentation is constant upon changing drums. This depth was tuned manually to give sufficient contact for reliable tactile signals while minimizing wear on the abrasive surface of the rotating drums}. Each drum is rotated over a set of $N_{\rm{speeds}}$ = 10 linear speeds, linearly spaced from 10 to 100\,mms\textsuperscript{-1}, giving a total of $N_{\rm{tex}}N_{\rm{speeds}}$ = 130 stimulus conditions. During each experimental run, the TacTip is stimulated for 65\,sec per stimulus condition whilst simultaneously recording TacTip video and microphone data. The initial and final 2.5\,sec of each recording are thrown away to keep only the constant-speed stimulation.  

\subsubsection{Tactile Image Data}
The TacTip camera output for each stimulus condition is truncated to 5000 frames (55\,sec) to ensure an equal number of samples per class. Samples are generated from individual frames. To examine the capacity for these methods to enable speed-invariant texture classification, tactile image data is separated into training, validation and testing sets using a {10-fold leave-one-speed-out cross-validation} procedure, where in each of the 10 individual sets a different stimulus speed is `held out'. Data from the held out speed is randomly split (50:50) to produce validation and test sets and the remaining 9 speeds used for training.\\
\linebreak
\noindent {\bf Spatial encoding:}
Samples are constructed from individual frames, {where $N_{\rm{train}}\!=\!45000$, $N_{\rm{val}}\!=\!2500$ and $N_{\rm{test}}\!=\!2500$} samples per class respectively. Over all textured stimuli, $N_{\rm{tex}}N_{\rm{train}}\!=\!585000$, $N_{\rm{tex}}N_{\rm{val}}\!=\!32500$ and $N_{\rm{tex}}N_{\rm{test}}\!=\!32500$ samples respectively. Tactile spatial encoding arises from the spatial structure within the image frames captured by the camera. \hfill \hfill \\
\linebreak
\noindent {\bf Spatio-temporal encoding:}
Samples are constructed from sets of 10 adjacent tactile images, yielding a total of 500 samples per stimulus condition (in analogy with human perception, this is a 0.1\,sec encoding window). Thus, $N_{\rm{tex}}N_{\rm{train}}$ = 58500, $N_{\rm{tex}}N_{\rm{val}}$ = 3250 and $N_{\rm{tex}}N_{\rm{test}}$ = 3250 samples respectively. Each sample is now a 640$\times$480$\times$10 3D array, where the decoding model can extract features contained within this full 3D spatio-temporal structure.

With a finite amount of data, there is a trade-off between encoding window length and the number of samples. \textcolor{black}{We experimented with different duration of encoding windows and found that a window of 0.1\,sec (10 time samples) provided sufficient numbers of samples from our training data whilst also enabling the validation accuracy to reach a reasonable value.} 

\subsubsection{Microphone Data:}

\indent {\bf Temporal encoding:}
Overlapping samples of 1\,sec with stride 0.5\,sec are generated from sets of 88200 audio data points, giving 58 samples per stimulus condition. As with the tactile image data, microphone data is separated into train, validation and test sets using a similar {10-fold leave-one-speed-out cross-validation} procedure. In each set, a different stimulus speed is `held out' and randomly split 29:30 for validation and test sets. The remaining 9 speeds are used for training, giving $N_{\rm{tex}}N_{\rm{train}}\!=\!6903$, $N_{\rm{tex}}N_{\rm{val}}\!=\!377$ and $N_{\rm{tex}}N_{\rm{test}}\!=\!390$ samples respectively. These data samples permit only a purely {temporal encoding} over the 1D array of temporal data.

\subsection{Feature Extraction}
\label{sec02:05}

The methods below for generating artificial SA-I and RA-I afferents are shown in a related paper to model many aspects of firing rate codes of their natural counterparts, \textit{e.g.} adaption rates, spatial modulation and sensitivity to edges, and have been shown to provide biologically plausible population codes for grating orientation discrimination \cite{pestell_gratings}. Furthermore, the RA-I model is similar to other transduction models where the first derivative of pressure is used as the primary input to a biological neuron model \cite{lee,bensmaia6,kim2}. 

Likewise, the processing of the vibrotactile channel is based on the prevailing theory of human texture perception that proposes the observed invariance to scanning speed is based on mechanisms  analogous to those that give timbre invariance in hearing \cite{saal2}. The harmonic structure of induced vibrations and resulting neuron firing is consistent across scanning speeds \cite{yau,boundy-singer}, with somatosensory cortex extracting these textural properties from the harmonic structure of neural activity. 

\subsubsection{Artificial SA-I Afferents}
\label{sec02:05:01}

SA afferents refer to a class of afferents with slow adaptation rates. That is, they respond to sustained pressure or deformation of the skin as their associated mechanoreceptors (Merkel cells) fire tonically. Type-I implies that the afferents densely innervate skin regions and exhibit small receptive fields with well defined borders \cite{johansson2}. They are therefore associated with high spatial acuity.     

Artificial SA-I activity for each channel is modelled by Euclidean distance of a marker from its at-rest position. Considering a sustained stimulus, the deformation of the tip will remain consistent and therefore so will the positions of the markers. In practice, marker positions are extracted from tactile image data with a simple blob detection algorithm implemented using OpenCV in Python. The artificial SA-I response at frame $i$ for marker $n$, ${\rm SA}_{n,i}$, is computed as the Euclidean distance between the marker positions $(x_n,y_n)$ and an initial undisturbed at-rest frame ($i=0$):
\begin{equation}
\centering
{\rm SA}_{n,i} = \sqrt{(x_{n,i}-x_{n,0})^{2}+(y_{n,i}-y_{n,0})^{2}}.
\label{SA-fire}
\end{equation}
Thus the artificial SA-I spatial encoding and spatio-temporal encoding samples are of dimensions 19$\times$19 and 19$\times$19$\times$10 respectively.

\subsubsection{Artificial RA-I Afferents}
\label{sec02:05:02}

In contrast, RA-I afferents are rapidly adapting, meaning their firing tends to decrease rapidly when subject to sustained stimulus; they also densely innervate the skin \cite{johansson3} and have small receptive fields that can be again associate with physical movement of the TacTip markers. It is believed that these afferents are particularly sensitive to the velocity of the skin within the receptive field \cite{johansson2,knibestol2} and thus tend to fire whenever skin is moving. We model this behaviour as marker speed, which is inherently transient, \textit{i.e.} it will be zero when the stimulus is sustained and positive when the stimulus changes. 

Artificial RA-I activity is derived from the same tactile image data as the SA-I activity. The response at frame $i$ for marker $n$, ${\rm RA}_{n,i}$, is computed as the \textcolor{black}{absolute} difference between the artificial SA-I responses on adjacent frames, \textcolor{black}{representing a radial speed of marker motion relative to the marker positions in the initial at-rest frame:
\begin{equation}
\textcolor{black}{
\centering
{\rm RA}_{n,i} = |{\rm SA}_{n,i} - {\rm SA}_{n,i-1}|.
}
\label{FA-fire}
\end{equation}}
As with artificial SA-I samples, RA-I spatial encoding and spatio-temporal encoding samples are of dimensions 19$\times$19 and 19$\times$19$\times$10 respectively.

\textcolor{black}{One should recognize that Equations~(\ref{SA-fire},\ref{FA-fire}) are simplifications of the afferent activity and may not encode some important skin dynamics; for example, \textcolor{black}{neither the sign nor circumferential component} of the marker velocity is represented in the artificial RA-I activity. Our intent here is to provide a parsimonious representation of afferent activity from simple transformations of the optical tactile output, as a foundation for improved models in the future.} 

\subsubsection{Vibrotactile Channel}
\label{sec02:05:03}

\begin{figure*}[t]
	\begin{center}
        \includegraphics[width=\textwidth, trim=0cm 1.5cm 0cm 1.5cm]{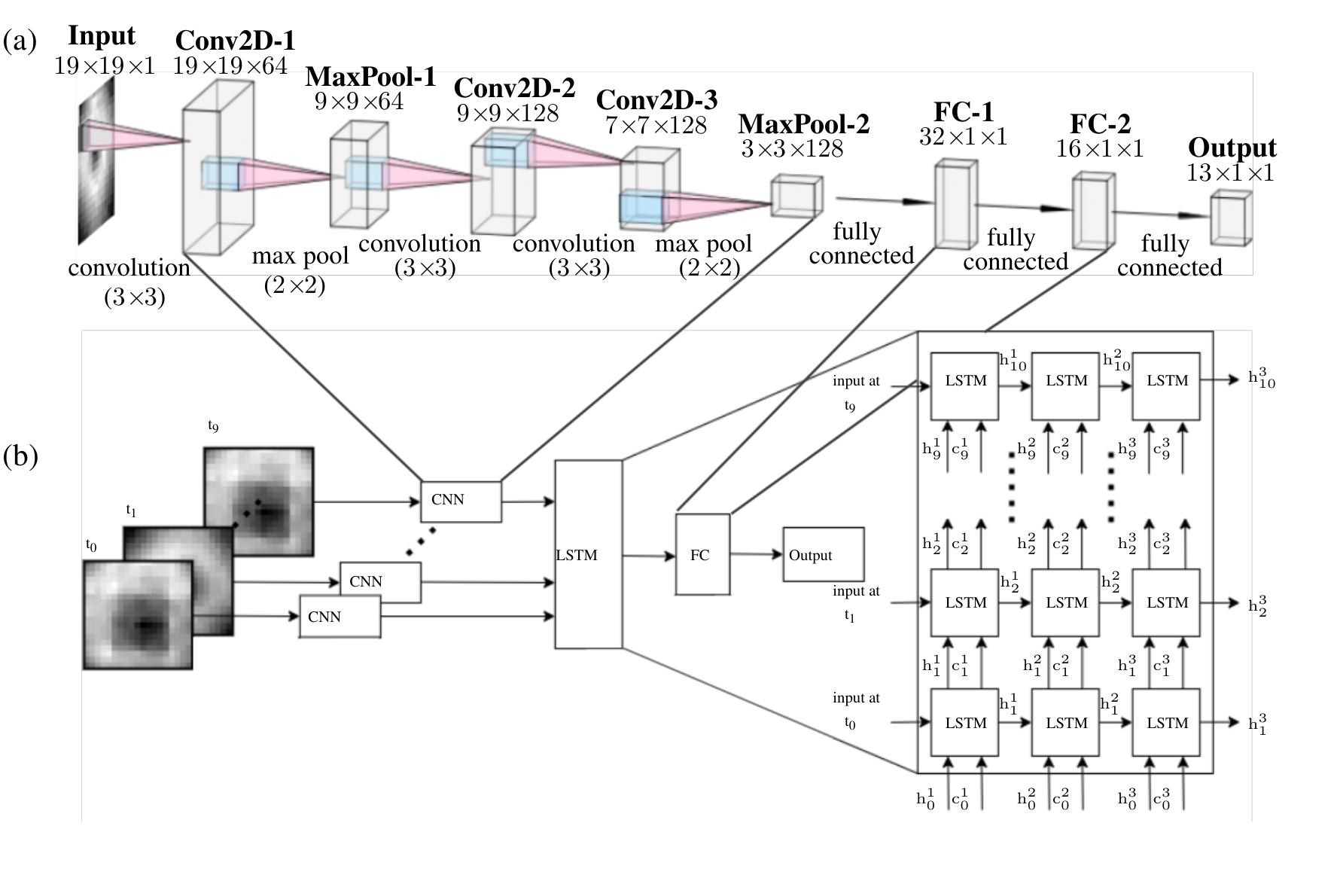}
	\vspace{-.5em}
	\caption{Spatial and spatio-temporal encoding and classification models. (a) Spatial encoding with a 2D-CNN architecture applied to the tactile images. (b) Spatio-temporal encoding with a CNN$\times$LSTM architecture applied to a time-sequence of tactile images that are combined with an LSTM block between the convolutional (Conv) and fully connected (FC) layers.)} 
	\label{cnn}
	\end{center}
	\vspace{1em}
	\begin{center}
		\includegraphics[width=\textwidth, trim=0cm 1.5cm 0cm 1cm]{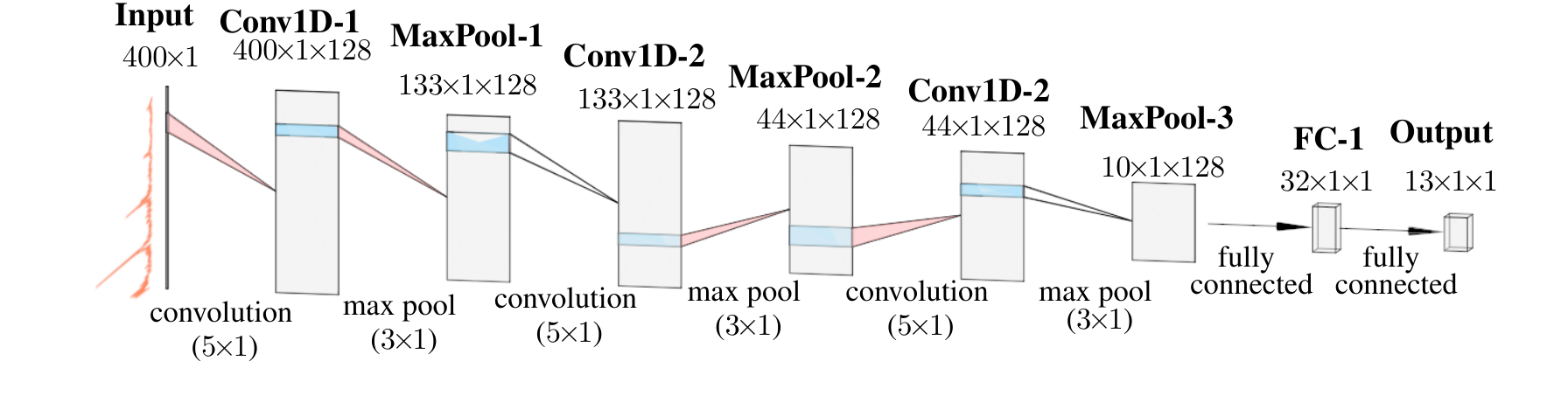}
	\vspace{1em}
	\caption{Temporal encoding model for vibration channel. Inputs are discrete FFTs of vibration data samples fed through a 1D-CNN architecture and fully connected layers for classification.)} 
	\label{cnn_audio}
	\end{center}
	\vspace{-1em}
\end{figure*}

A second rapidly adapating population, the RA-II afferents, have innervation densities considerably lower than that of type-I afferents \cite{westling}. Each RA-II afferent is terminated by a single Pacinian corpuscle, which are sparsely located deep in the subcutis with large receptive fields having obscure boundaries. They are sensitive primarily to high frequency vibrations (50-500\,Hz) with extremely low-amplitude thresholds (1\,$\mu$m) in the range of 100-300\,Hz \cite{johansson6}.

In design, the vibrotactile channel is analogous to the form of the natural RA-II channel in two ways: firstly, the microphone innervates the TacTip sparsely relative to markers; secondly, it is located deep within the soft silicone gel analogous to the dermis. The information acquired by the elecret microphone is also similar to the dynamics of the natural RA-II channel in being: (i) responsive to physical vibrations (acoustics), which are transient by definition; (ii) having an operating frequency range \textcolor{black}{10\,Hz-20\,kHz (sampled at 44.1\,kHz)} that \textcolor{black}{contains that of the Pacinian system in its lower range}; and (iii)~exhibiting extreme sensitivity to low-amplitude vibrations. \textcolor{black}{Note also that the front sensitive area of the microphone is embedded within the supporting visco-elastic gel of the tactile sensor, which appears to have attenuated frequencies above $\sim$150\,Hz (e.g. Figure~\ref{FFTs_augmented}). Hence, the effective range of the embedded microphone is more similar to that of natural RA-II afferents than an auditory range.}

The method of feature extraction for the artificial vibrotactile channel is based on extracting its harmonic structure. From each sample of raw audio data in the time domain, a 1D feature vector of amplitudes at discrete frequencies is constructed with a fast Fourier transform (FFT). \textcolor{black}{These feature vectors are truncated to a maximum frequency of 200\,Hz, corresponding to a discrete feature vector size of 200$\times$1. The amplitudes are scaled to fall between 0 and 1 with a division by the largest amplitude across the entire training data set.}

\subsection{Tactile Encoding Models}
\label{sec02:06}

All tactile encoding models are trained in the same way using Keras with TensorFlow backend. Training data is fed through the networks in batches of 64 samples. The entire set is propagated for a maximum of 150 epochs. An Adaptive Momentum Estimation optimiser (ADAM) on a categorical cross-entropy loss is used for updating the weights after each batch. Training may stop early if validation accuracy plateaus with the use of a patience factor of 30 epochs. For the spatial, spatio-temporal and temporal encoding/decoding methods, a separate model is trained for each of the 10 held-out speeds.

Overall, there are 10 spatial encoding SA-I and \mbox{RA-I} texture classification models, one per speed~\textit{{v}}: SA-SE-\textit{{v}} and RA-SE-\textit{{v}} r\mbox{espe}ctively, and likewise 10 spatio-temporal models, SA-STE-{\textit{v}} and \mbox{RA-STE-\textit{{v}}}. There are also 10 temporal encoding vibrotactile texture classification models, vib-TE-\textit{{v}}.

\subsubsection{SA-I and RA-I Afferents - Spatial Encoding Models}
\label{sec02:06:01}

Spatial decoding models of artificial SA-I and RA-I afferents are constructed with a 2D convolutional neural network (schematic shown in Figure \ref{cnn}A). This network permits 19$\times$19 tactile images as input and outputs a 13$\times$1 vector with each element corresponding to an individual texture class. All hidden layers use ReLU activation functions and the output layer uses softmax activations on each neuron. Regularization techniques include drop-outs of 0.4, 0.2 and 0.2 prior to layers FC-1, FC-2 and the Output respectively, and L2 regularisation (factor 0.005) on each dense layer. Batch normalisation is used after each convolutional layer. This architecture was chosen by manually tuning network hyper-parameters \textcolor{black}{for good validation performance over all textures using a single set of hyper-parameters for all models for each of the leave-one-out speeds. In practice, this tuning was straightforward and many similar architectures would have worked well.}

\subsubsection{SA-I and RA-I Afferents - Spatio-temporal Encoding Models}
\label{sec02:06:02}

Spatio-temporal decoding models of artificial SA-I and RA-I afferents are constructed with a convolutional-LSTM (ConvLSTM) built over the 2D convolutional network described above (schematic shown in Figure \ref{cnn}B). 

In essence, a frame at each timestamp, $t_i$, of a sample is passed through a feature detector `CNN' taken directly from that used in the spatial decoding model (Figure \ref{cnn}A; layers Conv2D-1 to MaxPool-2). The output feature maps are flattened before being passed to a multi-layer LSTM unit, consisting of 3 layers with 10 LSTM blocks each. The output of the LSTM unit is passed to a fully connected unit `FC' also taken directly from the spatial decoding model (Figure \ref{cnn}A: layers FC-1 to FC-2). An output layer, consisting of 13 neurons, provides a prediction for each class.

Reusing the feature detection and fully-connected parts of the spatial decoding model provides a more controlled comparison between encoding mechanisms; \textit{i.e.} differences in performance between static and dynamic touch can more easily be attributed to the availability of temporal features. The LSTM hyper-parameters were chosen by manual tuning \textcolor{black}{for good validation performance over all textures using a single set of hyper-parameters for all models.}

\subsubsection{Vibrotactile Channel - Temporal Encoding Models}
\label{sec02:06:03}

Temporal decoding models of the vibrotactile channel are constructed with a 1D convolutional neural network (schematic shown in Figure \ref{cnn_audio}). This network takes 200$\times$1 harmonic feature vectors as input and outputs a 13$\times$1 vector corresponding the texture classes. All hidden layers use ReLU activation functions and the output layer uses softmax activations on each neuron. Regularization techniques included a drop-out of 0.4 prior to layer FC-1 and L2 regularisation (factor 0.005) on layer \mbox{FC-1}. Batch normalisation is used after each convolutional layer. Again, this architecture was chosen by manually tuning hyper-parameters \textcolor{black}{for good validation performance over all textures for all models}. 

\begin{figure}[t]
 	\begin{center}
        \includegraphics[width=\textwidth]{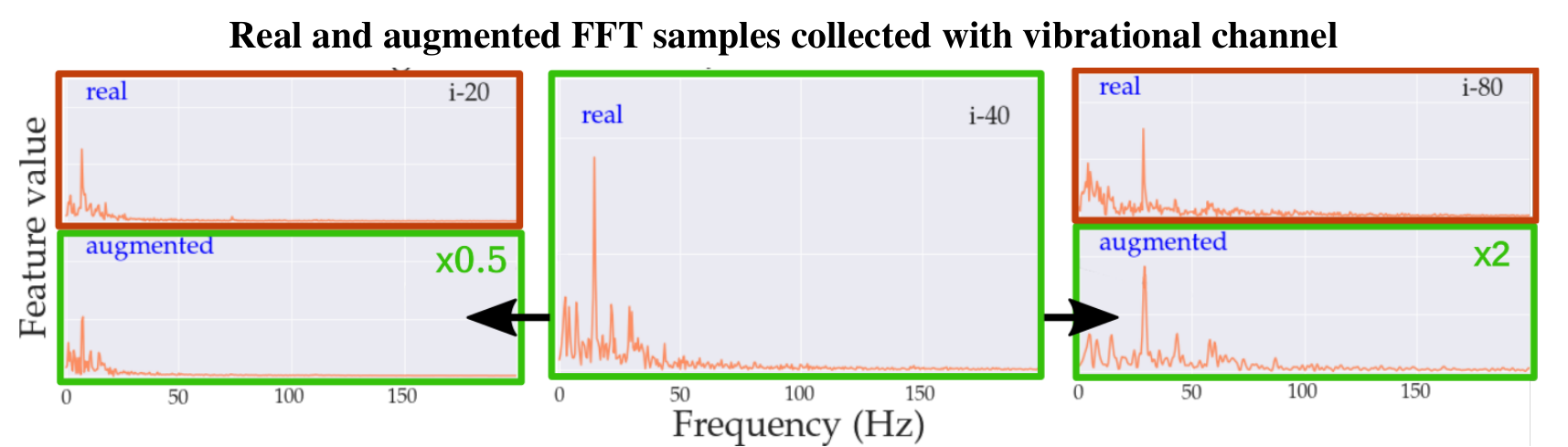}
 	\end{center}
 	\vspace{-1.5em}
	\caption{Real and augmented vibrotactile samples from distinct speeds on stimulus \textit{\textbf{i}}. \textit{Centre}: original data collected at 40\,mms\textsuperscript{-1} used for augmentation. \textit{Left}: original data collected at 20\,mms above augmented data compressed by a stretching factor of 0.5. \textit{Right}: original data collected at 80\,mms\textsuperscript{-1} above augmented data stretched by a factor of 2.}
	\label{FFTs_augmented}
\end{figure}

A data augmentation procedure is implemented to improve the accuracy of these temporal encoding/decoding methods \textcolor{black}{and encourage generalization across speeds and textures.} Informed by the hypothesis that a speed-invariant harmonic structure of induced vibration is a viable cue for textural properties \cite{saal2,boundy-singer}, we propose that data collected at different speeds can be simulated by `stretching' and `compressing' the FFT samples in the frequency domain \textcolor{black}{(Figure~\ref{FFTs_augmented})}. For each original training sample, six augmented samples are generated: three by stretching in the frequency domain by a factor uniformly sampled between \textcolor{black}{scaling factors of~[1,2] to represent higher speeds} and the remaining three compressed by a factor uniformly sampled between \textcolor{black}{scaling factors of~[0.5,1] to represent lower speeds}. 

\textcolor{black}{Overall, $6\times N_{\rm{train}}\!=\!714$ augmented training samples per texture and speed are generated, where $N_{\rm{train}} = 119$ is the original number of training samples per texture class and speed. For each of the 10 temporal encoding models, $6\times 9\times13\times N_{\rm train}=83,538$ augmented training samples are available over all 13 textures trained in the leave-one-speed-out classification model.}

\begin{table}[b]
\vspace{0em}
	\centering
	\caption{Comparative accuracies of each encoding mechanism, averaged across hold-out speeds \textcolor{black}{(for comparison, the random chance over 13 textures is about 8\%).}}
	\begin{tabular}{cccc}
	\textbf{model} & \textbf{afferent} & \textbf{encoding} & \textbf{accuracy} \\
	\hline
	SA-SE & SA-I & spatial & 46\% \\
	RA-SE & RA-I & spatial & 53\% \\
	SA-STE & SA-I & spatio-temporal & 46\% \\
	RA-STE & RA-I & spatio-temporal & 70\% \\
	vib-TE (real data) & RA-II/vibrotactile & temporal & 50\% \\
	vib-TE (augmented) & RA-II/vibrotactile & temporal & 90\% \\
	\end{tabular}
	\label{accs}
\end{table}

\pagebreak

\begin{figure}[t!]
    \begin{center}
    \includegraphics[width=\textwidth]{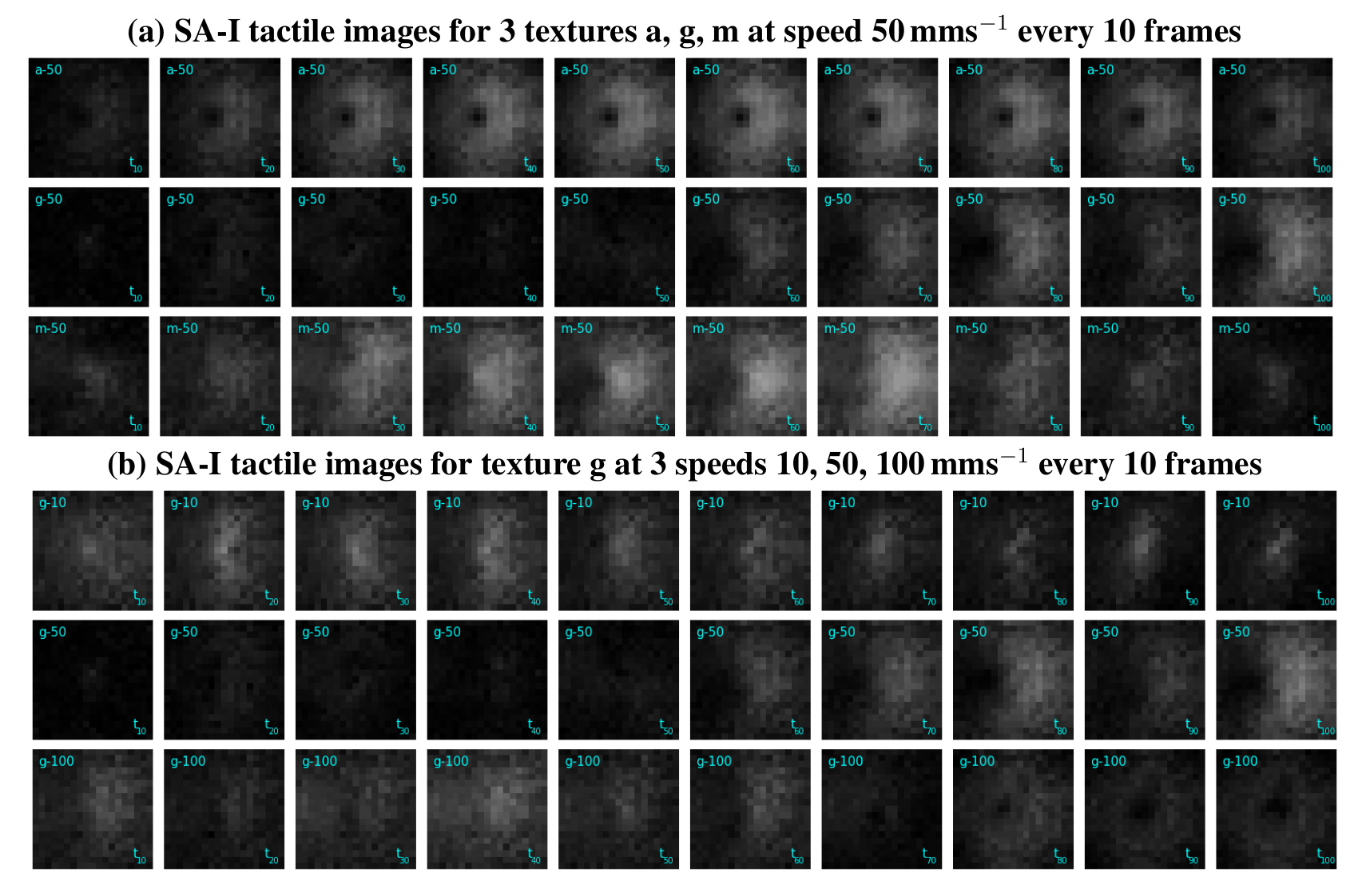}
    \end{center}
	\vspace{-1.5em}
	\caption{Tactile images of artificial SA-I responses according to textured stimulus and speed (denoted by the letter and number; see Table 2). (a) Examples from speed 50\,mms\textsuperscript{-1} on the smooth {\textbf{a}}, middle-coarseness {\textbf{g}} and coarsest stimuli {\textbf{m}}. (b) Examples from the middle-coarseness {\textbf{g}} stimulus at the minimum 10\,mms$^{-1}$, middle 50\,mms$^{-1}$ and maximum 100\,mms$^{-1}$ speeds. For each case, 10 frames spaced every 10 time samples ($t_{10}$, $t_{20}$ to $t_{100}$) are shown to indicate the response variance. \textcolor{black}{Note that the texture pattern is not visible in the tactile image, even for the coarsest stimulus {\textbf{m}}. We attribute this effect to the stimulus motion and the relaxation time of the skin, rather than the image resolution.}}
	\label{SA_tactile_images_slides}
\end{figure}

\section{Results}
\label{sec03}

\subsection{Artificial SA-I Afferents}
\label{sec03:01}

\subsubsection{Visual Inspection of Data}
Example SA-I tactile images are shown for the smoothest {\textbf{a}}, middle {\textbf{g}} and coarsest {\textbf{m}} textured stimuli and at the slowest, middle and fastest speeds \mbox{(10-100\,mms\textsuperscript{-1})} (Figures \ref{SA_tactile_images_slides}). The 10 example tactile images for each texture were collected at the speed indicated on the image label  {\em e.g.} {\textbf{g-10}} is for texture \textbf{g} at 10\,mms\textsuperscript{-1}.

From visual inspection, we could not observe any features in the SA-I tactile images that robustly signified texture class. For example, images \textbf{m-20} and \textbf{g-40} in Figure \ref{SA_tactile_images_slides}A for different texture classes look similar, whilst there is a considerable variation with speed within both texture classes {\textbf{m}} and {\textbf{g}}. 

To eliminate the possibility that the variation in Figure~\ref{SA_tactile_images_slides}A is due to stochastic variation in time with contacting each texture, Figure \ref{SA_tactile_images_slides}B shows tactile images of artificial SA-I on stimulus class \textit{\textbf{g}} (class with the median bump spacing). Rows are separated according to speed (10, 50 and 100\,mms\textsuperscript{-1}) and consist of time series of 10 consecutive frames (increasing times $t_1$ to $t_{10}$). Tactile images appear similar in time (rows) and dissimilar in speed (columns). 

Considering Figures \ref{SA_tactile_images_slides}A,B in combination, we conclude that the spatial modulation of artificial SA-I firing over the TacTip is highly dependent on both texture and speed. The dependence is such that changes in speed can compensate changes in texture (and vice-versa).  

Note also that many of the SA-I tactile images exhibited asymmetry. For example, a light crescent shape to the left or right of the image (\textit{e.g.} \textbf{g-40}, \textbf{g-30}, \textbf{m-20}), suggesting that when stimulated via sliding, the TacTip's deformation was dominated by shearing effects.

\begin{figure}[t!]
    \begin{center}
           \includegraphics[width=\textwidth]{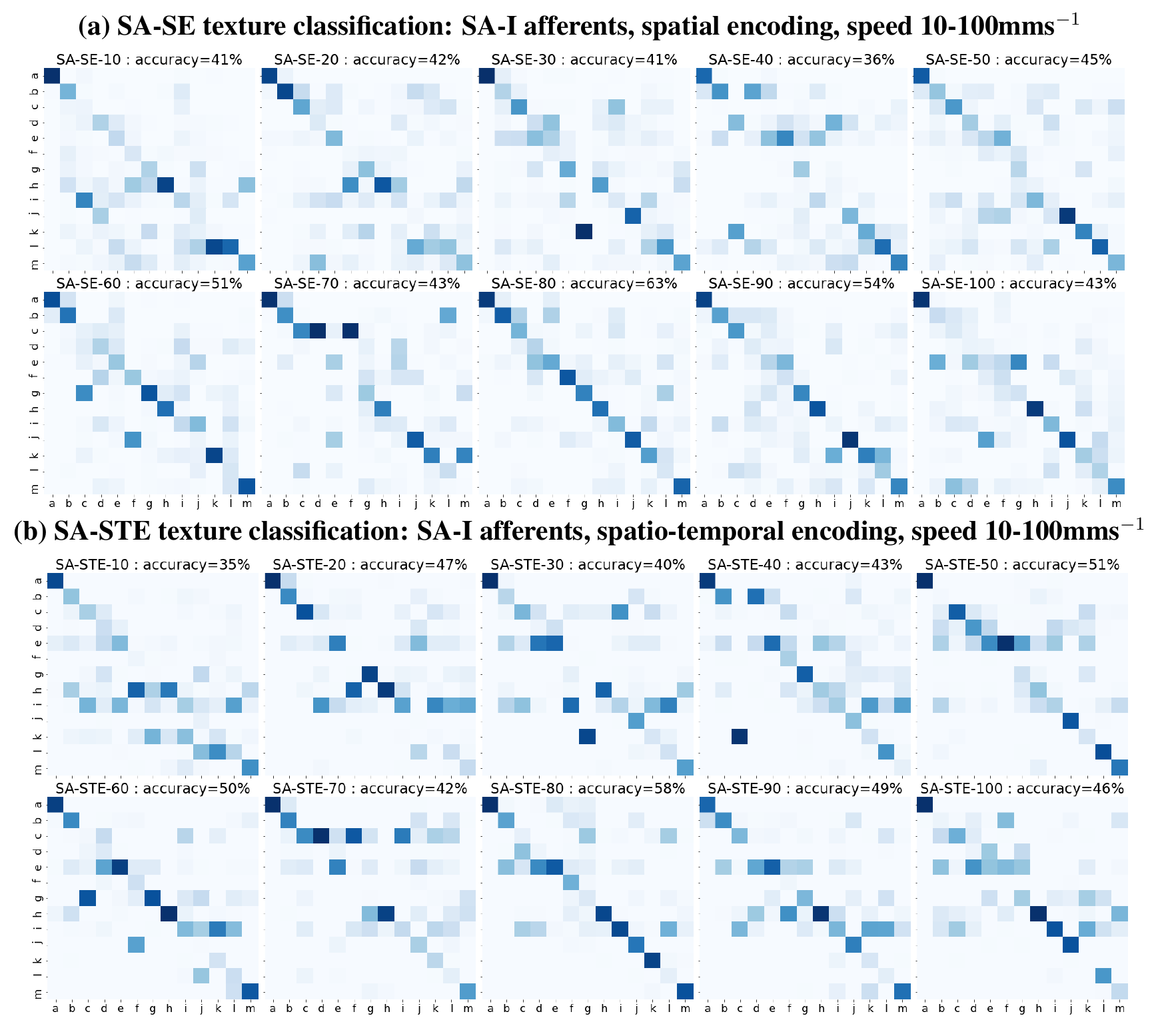}
    \end{center}
	\vspace{-2em}
	\caption{Confusion matrices for texture classification with artificial SA-I tactile afferents. (a) Spatial encoding model (SA-SE-v) tested on the 10 hold-out speeds $v$ from 10 to 100\,mms$^{-1}$. (b) Spatio-temporal encoding model (SA-STE-v) tested on the same speeds.} 
	\label{SA_slide-CM_l1o}
\end{figure}

\subsubsection{Texture Classification}

Accuracies of all texture classification models over artificial SA-I, RA-I and vibrotactile afferents averaged across the hold-out speeds are reported in Table~\ref{accs}. Here, we focus on the SA-I afferents, and show the  corresponding confusion matrices using temporal (TE) and spatio-temporal encoding (STE) schemes in Figures \ref{SA_slide-CM_l1o}A,B respectively.

For both the spatial and spatio-temporal encoding models (SA-SE, SA-STE), each model performed better than chance suggesting that there is coherence within texture classes in terms of speed-invariant spatial features (10 panels over \textit{v} in each of Figures~\ref{SA_slide-CM_l1o}A,B). The overall accuracy averaged across all 10 models for the distinct speeds was $\sim$46\% in both cases (Table~\ref{accs}), which were the lowest of all models considered. 

For all hold-out speeds, near-perfect predictions were made for stimulus class \textit{\textbf{a}} for a completely smooth texture (top-left of all panels in Figures~\ref{SA_slide-CM_l1o}A,B). Overall, there is a general trend towards better performance at either end of the range of stimuli, which  may be because textures at the extremes of the roughness scale have fewer neighbouring classes, making prediction easier. In addition, the extremes may be more distinctive; for example, the completely smooth texture is easily recognizable.

The best-performing SA-I model was for spatial encoding with hold-out speed of 80\,mms\textsuperscript{-1} (SA-SE-80), attaining a classification accuracy of 63\%. There appears to be a trend between increasing hold-out speed and improved performance, \textit{i.e.} more salient data being created when the stimulation speed is increased. However, this relationship seems to be attenuated at the extremes of the speed range, which we attribute to the network being trained only on data collected either at faster or slower speeds for 10 and 100\,mms\textsuperscript{-1} respectively, leading to poorer generalization.

Overall, the spatial and spatio-temporal encoding models (SA-SE, SA-STE) performed similarly, suggesting that models based on SA-I afferents do not benefit from the addition of a temporal dimension; \textit{i.e.} artificial SA-I afferents do not encode texture temporally. 

\begin{figure}[t!]
    \begin{center}
        \includegraphics[width=\textwidth]{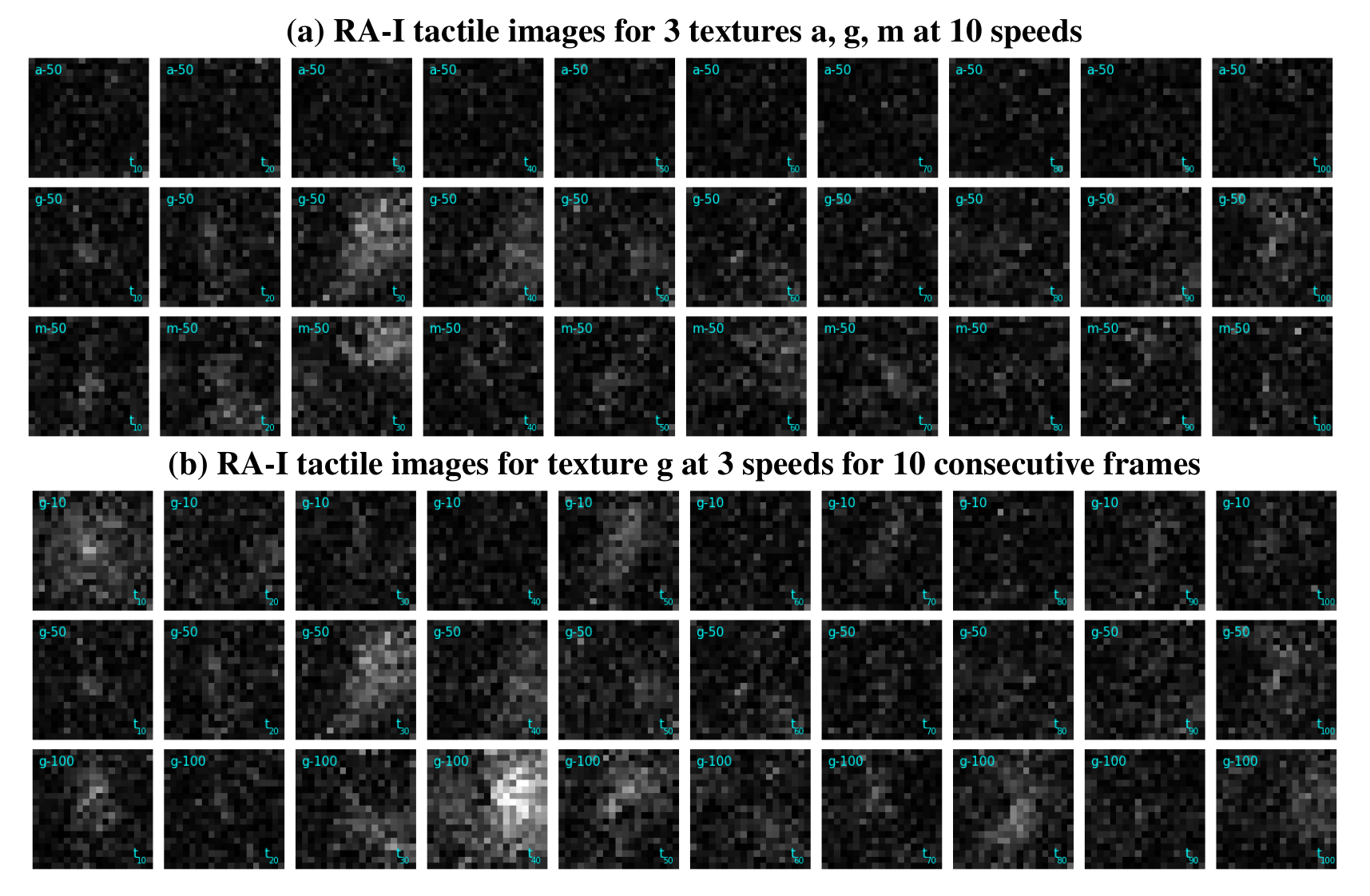}
    \end{center}
	\vspace{-2em}
	\caption{Tactile images of artificial RA-I responses according to textured stimulus and speed (denoted by the letter and number; see Table 2). (a) Example from each speed (10-100\,mms\textsuperscript{-1}) on the smooth {\textbf{a}}, middle-coarseness {\textbf{g}} and coarsest stimulus {\textbf{m}}. (b) Series of 10 consecutive time-samples ($t_1$ to $t_{10}$) collected at speeds 10, 50 and 100\,mms\textsuperscript{-1} on stimulus \textit{\textbf{g}}. \textcolor{black}{Note, similarly to Figure 8, no spatial pattern is visible in the tactile image, even on the coarsest stimulus.}}
	\label{RA_tactile_images_slides}
\end{figure}

\subsection{Artificial RA-I Afferents}
\label{sec03:02}

\subsubsection{Visual Inspection of Data}

Example RA-I tactile images are shown for the smoothest ({\textbf{a}}), middle ({\textbf{g}}) and coarsest ({\textbf{m}}) textured stimuli (Figure \ref{RA_tactile_images_slides}A) made from frames corresponding to those for the SA-I tactile images shown in Figure \ref{SA_tactile_images_slides}A. Likewise, Figure \ref{RA_tactile_images_slides}B shows example RA-I tactile images just for the middle stimulus class ({\textbf{g}}) for 3 speeds (columns: $v=$10, 50 and 100\,mms\textsuperscript{-1}) and 10 consecutive frames (rows: times $t_1$ to $t_{10}$). 

There is less structure within these artificial RA-I tactile images compared with their SA-I counterparts; for example, the geometry of the contact region is no longer as apparent in Figures~\ref{RA_tactile_images_slides}A,B. This difference in apparent structure is likely a consequence of the way the samples were constructed: each RA-I tactile image is in effect constructed from the difference in adjacent SA-I tactile images, which look similar (Figure~\ref{SA_tactile_images_slides}B). However, even though there is less geometric structure, the information may still be informative about the texture, as will be examined below.

As with the artificial SA-I tactile images, RA-I images do not exhibit any clear visual features which robustly signify texture class (Figure \ref{RA_tactile_images_slides}A), except for an apparent difference in intensity between the textured ({\textbf{g}},{\textbf{m}}) and smooth ({\textbf{a}}) classes. Contrastingly, however, the artificial RA-I tactile images do not show the same degree of consistency between adjacent images (Figure \ref{RA_tactile_images_slides}B), indicating there may be information about texture encoded in the time-sequence of these tactile images.

\begin{figure}[t!]
    \begin{center}
        \includegraphics[width=\textwidth]{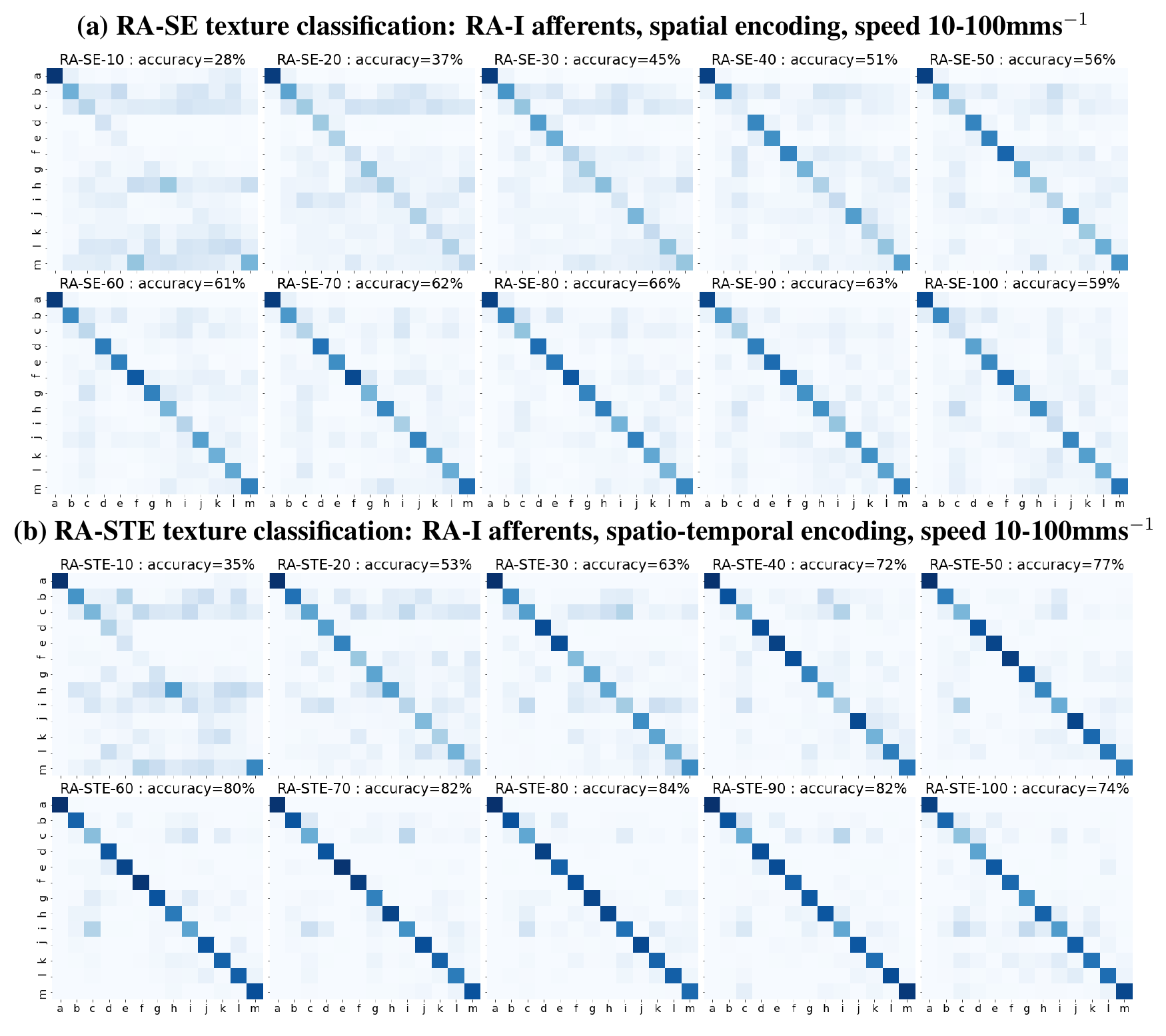}
    \end{center}
	\vspace{-2em}
	\caption{Confusion matrices for the accuracy of texture classification with artificial RA-I tactile afferents. (a) Spatial encoding model (RA-SE-v) tested on the 10 hold-out speeds $v$ from 10 to 100\,mms$^{-1}$. (b) Spatio-temporal encoding model (RA-STE-v) tested on the same speeds.} 
	\label{FA_slide-CM_l1o}
\end{figure}

\subsubsection{Texture Classification}

Texture classification with the artificial RA-I afferents is now considered by showing the corresponding confusion matrices using spatial (SE) and spatio-temporal encoding (STE) models in Figures \ref{FA_slide-CM_l1o}A,B respectively. Accuracies averaged across all hold-out speeds are reported in Table~\ref{accs}.

Both the spatial and spatio-temporal encoding models (RA-SE, RA-STE) tended to perform better than the corresponding SA-SE models for the SA-I afferents, as seen from confusion matrices of RA-I models (Figures \ref{FA_slide-CM_l1o}A,B) compared to SA-I models (Figures \ref{SA_slide-CM_l1o}A,B). However, the average accuracy of spatial encoding RA-SE models across held-out speeds (53\%) was only slightly higher than that of the SA-SE models (Table~\ref{accs}). This difference is due partly to the poorer accuracy of the RA-SE-10 model (28\%) than the SA-SE-10 model (40\%). Artificial RA-I afferent performance also exhibited improved performance for higher hold-out speeds, in a similar trend to that seen for artificial SA-I performance. 

Overall, the spatio-temporal models (RA-STE) performed best with a mean accuracy of 70\% (Table~\ref{accs}) and the largest individual model accuracies (\textcolor{black}{RA-STE-\textit{{v}}}, Figure \ref{FA_slide-CM_l1o}B). This suggests that artificial RA-I afferents can encode texture in the spatio-temporal modulation of their responses; \textit{i.e.}, there is benefit to extracting spatio-temporal over purely spatial features. This result contrasts with SA-I artificial afferents, a result predicted from inspection of the tactile images (Figures \ref{RA_tactile_images_slides}B vs \ref{SA_tactile_images_slides}B), where more variation was seen within a single dynamic sample (row) of artificial RA-I responses compared to artificial SA-I responses.

That artificial RA-I afferents were generally better predictors of texture than SA-I afferents when using a held out speed, suggests that artificial RA-I afferents offer better generalisation to previously unobserved speeds. Equivalently, SA-I models tend to over-fit, responding to spatial structures that vary with speed rather than speed-invariant features. This hypothesis is supported by comparing Figurse \ref{SA_tactile_images_slides}B,\ref{RA_tactile_images_slides}B for the SA-I and RA-I responses: within a given speed there appears to be little coherence in terms of spatial structure for RA-I firing, unlike with SA-I tactile images. Thus, RA-I data makes the model less likely to learn undesirable features; \textit{i.e.} the data is naturally regularised.   

\subsection{\textcolor{black}{Artificial RA-II} Afferents}
\label{sec03:03}

\subsubsection{Visual Inspection of Data}

\begin{figure}[p]
	\begin{center}
        \includegraphics[width=\textwidth, trim=0cm 1cm 0cm 1cm]{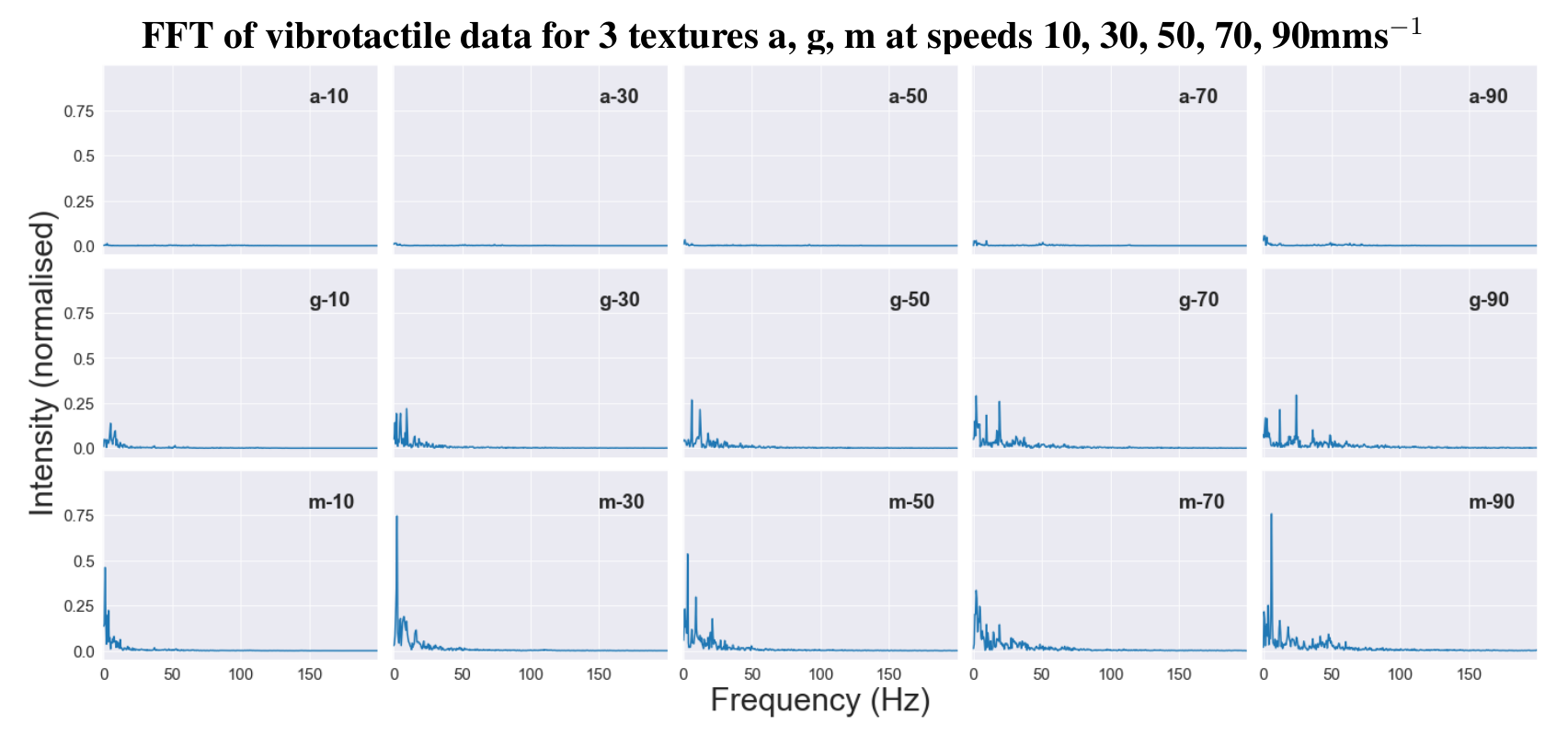}
	\end{center}
	\caption{Confusion matrices of dynamic texture prediction using \textit{leave-one-speed-out cross validation} with models trained on augmented vibration channel data. Each matrix refers to a different model trained and tested with the held-out speed on its individual title.}
 	\label{FFTs}
 	\vspace{-1em}
	\begin{center}
        \includegraphics[width=\textwidth]{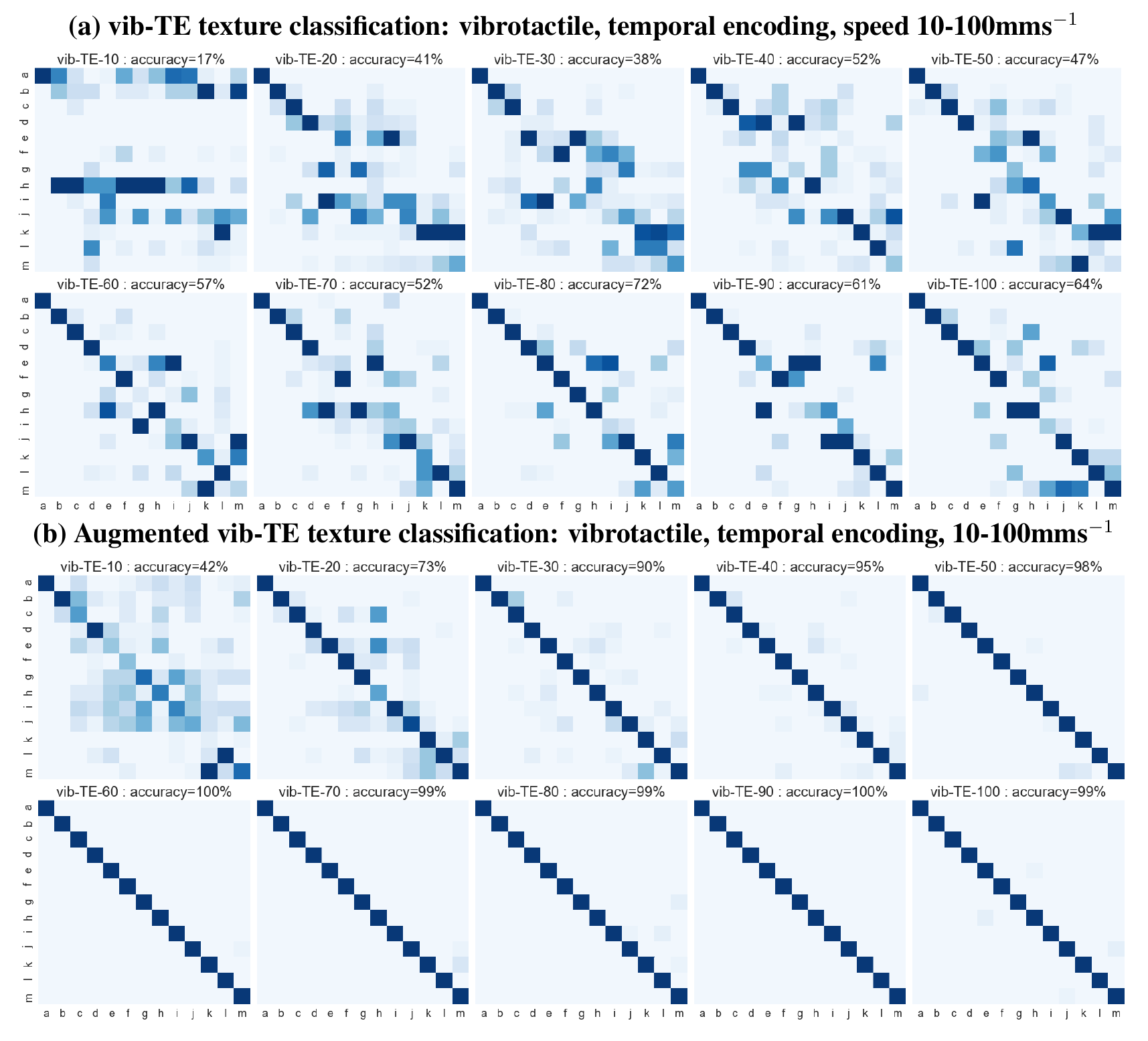}
	\end{center}
	\vspace{-2.5em}
	\caption{Confusion matrices of dynamic texture prediction using \textit{leave-one-speed-out cross validation} with models trained on (a) non-augmented and (b) augmented vibration channel data. Each matrix refers to a different model trained and tested with the held-out speed on its individual title.}
	\label{CM-l10_augmented}
\end{figure}

Examples of vibrotactile channel samples are generated from the harmonic structure of induced vibrations; \textit{i.e.} FFT shape (examples in Figure~\ref{FFTs}). Visual inspection suggests that the chosen resolution \textcolor{black}{(200 features at 1\,Hz resolution from 1\,sec of time-series data) provides sufficient detail to visually distinguish textures based on the shape of their FFTs up to 200\,Hz}, above which there was very little spectral power in all cases. \textcolor{black}{Consequently, there is a 1\,sec processing delay for this channel, compared with just 0.1\,sec used for the artificial SA-I and RA-I channels.} 

The hypothesis that the harmonic structure of induced vibrations is a viable cue for texture discrimination (Section~\ref{sec02:06:03}) is supported by the FFTs shown in Figure \ref{FFTs}: there is a consistent shape within each texture, with speed only affecting the scaling rather than structure. For example, at higher speeds the feature vectors are `stretched-out' with peak frequencies increased and at lower speeds the feature vectors are `compressed' with peak frequencies reduced. This structure led us to adopt an augmentation procedure to generate data samples for training, by `stretching' and `compressing' the original collected data accordingly. This was visualized with original and augmented vibration samples collected on stimulus {\textbf{i}} (Figure \ref{FFTs_augmented}). 

\subsubsection{Texture Classification}

Texture classification with the vibrotactile afferents is now considered by showing the corresponding confusion matrices using a temporal encoding (TE) model in Figure \ref{CM-l10_augmented} with either non-augmented or augmented data. For augmented data, the average accuracy across all hold-out speeds was 90\% (Table~\ref{accs}). Comparison models trained with just the original real data achieved an average accuracy of only 50\% and visibly poor confusion matrices. 

The improvement observed with augmented FFT samples provides strong evidence in support of the hypothesis that scanning speed scales frequency spectra whilst retaining harmonic structure. The additional training samples caused by augmentation have improved both the performance and generalization capabilities of the models. 

\textcolor{black}{Evidently, the augmented model performance is poorer at slower hold-out speeds (Figure~\ref{CM-l10_augmented}B, top-left panels). We attribute this to two effects: first, fewer data can be generated by the augmentation procedure if the data lies at the extreme of the speed range, e.g. for the 10\,mms\textsuperscript{-1} hold-out set, it is not possible to create similar augmented data by `stretching' data at lower speeds; second, physical effects that make textures at lower speeds less discriminable because their power spectra are compressed into narrower frequency range with fewer active features. Physical effects seem supported by the textures at higher speeds being accurately discriminated, even though again there is again fewer augmented samples available at the extreme of the speed range.}

Overall, the texture classification was near perfect for speeds of  30\,mms\textsuperscript{-1} or more using augmented data, with a mean accuracy over all speeds the largest (90\%) of all models, with the next best for the spatio-temporal encoding of RA-I artificial afferents (70\%; Table~\ref{accs}). We do caution an interpretation of the vibration modality being more informative about texture than tactile images, however, because no augmentation over speeds was used for the SA-I and RA-I models \textcolor{black}{(and it is non-obvious how such augmentation could be done). Moreover, there are other differences such as window size and sampling rate that differ between the SA-I/RA-I and RA-II modalities that also make a direct comparison less meaningful. The appropriate interpretation is that the augmented temporal encoding for the vibrotactile modality and spatio-temporal encodings for the RA-I modality both give good performance compared with other alternatives for those modalities.}  

\section{Discussion}

The artificial SA-I and RA-I afferents used in this study are shown in a related paper to offer viable models for their natural counterparts \cite{pestell_gratings}. Here we extended the perceptual test space to dynamic stimulation and enabled an additional temporal encoding dimension. In addition, we endowed \textcolor{black}{a soft biomimetic optical tactile sensor (the TacTip) with a novel and complementary modality for that sensor type: the \textit{vibrotactile channel}, which draws inspiration from RA-II afferents in human touch.}

\subsection{Frictional Cues Enable Salient Spatial Codes for \textcolor{black}{Texture Roughness}} Artificial SA-I and RA-I tactile images were dominated by global shearing as a consequence of frictional forces between the \textcolor{black}{dynamic} stimulus and the TacTip's skin, rather than spatial aspects seen in related work on static stimuli \cite{pestell_gratings}. \textcolor{black}{In particular, the texture pattern is not visible in the tactile image, which we attribute to the stimulus motion and relaxation time of the skin rather than the image resolution. Very light contacts with indentation less than the texture height (0.5\,mm) may have enabled greater use of local spatial features; however, our tactile sensing surface bulges slightly, so this would be difficult to test with the current design.} Even so, when leveraging only a spatial code, artificial SA-I afferents performed much better than chance (Figure \ref{SA_slide-CM_l1o}A) and artificial RA-I afferents provided accurate predictions of texture (Figure \ref{FA_slide-CM_l1o}A). \textcolor{black}{It is interesting that biological RA-I afferents also have the highest bandwidth for textural information, in being informative about the broadest range of textures, which is in accord with our results.} 

These results highlight a distinct difference between the present study, where sliding was used, and others using spatial encoding for texture classification in robotics that employed static or pressing stimulation \cite{li5,ojala,luo3,krizhevsky}. Although there are parallels in terms of the encoding mechanism, the tactile cue is distinct because of the difference in stimulation technique. Here, we have realised a frictional cue, visible as global shear in the tactile images, whereas static or pressing is likely to encode spatial cues only. Although it has been shown that humans may use spatial cues for coarse texture perception, \textit{e.g.} for subjective roughness judgements \cite{sathian,cascio,hollins} and discrimination \cite{hollins}, studies have shown that human surface discrimination may employ a `sticky-slippery' dimension which relies on frictional cues \cite{skedung}. This subjective sticky-slippery property has previously been utilised in robot texture discrimination for successful classification of real-world textures \cite{fishel}, although this was encoded in a frictional force measurement rather than spatially within tactile images.  

\subsection{Artificial RA-I Afferents Benefit from Temporal Encoding} Artificial RA-I afferents exhibited greater temporal variation than artificial SA-I afferents; for example, the difference in tactile images between adjacent frames was more significant (Figures \ref{RA_tactile_images_slides}B vs \ref{SA_tactile_images_slides}B). In human studies, it is understood that RA-I afferents are particularly responsive to vibrations (5-40\,Hz) \cite{muniak}, which are usually low amplitude compared with deformations associated with static touch. The heightened sensitivity to small-scale dynamic stimulation may explain why artificial \mbox{RA-I} afferents benefited from the availability of an additional temporal dimension (Figures \ref{RA_tactile_images_slides}A vs \ref{RA_tactile_images_slides}B), whereas artificial SA-I afferents did not (Figures \ref{SA_tactile_images_slides}A vs \ref{SA_tactile_images_slides}B). 

Neurophysiological experiments have shown that RA-I peripheral fibres respond more robustly than SA-I afferents to dynamic stimulation with a range of natural textures. Furthermore the frequency composition of the RA-I response reflected that of oscillations in skin induced by scanning \cite{weber2}, a finding which has informed many to believe that natural RA-I afferents are leveraged for mediating information about texture through a temporal representation of spike patterns \cite{weber2}. Here we demonstrated that by simply taking the temporal derivative across pairs of tactile images to construct artificial RA-I activity, temporal codes can be leveraged more effectively than on single tactile images.  

Whilst other studies have employed spatio-temporal encoding for robotic texture classification \cite{baishya,oddo,taunyazov}, to our knowledge ours is the first where the velocity component of raw tactile features have been used. Our finding that this encoding gave an effective discriminator of texture may have implications for these other studies, in particular to improve the effectiveness of temporal coding where its availability has so far provided little improvement \cite{yuan3}.  

\subsection{Harmonic Structure of Vibrotactile Channel Data Provides a Speed-Invariant Code for Texture Classification} It is widely believed that the natural Pacinian (RA-II) system is the primary tactile channel for mediating information about fine surface texture \cite{hollins3,weber2,bensmaia4}. We demonstrated highly accurate texture classification using a vibrotactile channel that is inspired by the natural RA-II channel (Figure \ref{CM-l10_augmented}B). For a given stimulus, the harmonic structure of induced vibrations (shape of the frequency spectra scaled by the scanning speed) is preserved across different scanning speeds; therefore, the harmonic structure of induced vibrations provides a speed-invariant code for texture discrimination. 

This finding is about the preservation of harmonic structure is interesting because it aligns with a compelling theory to explain speed-invariance of texture perception in humans. In particular, it has been proposed that cortical computations are capable of extracting harmonic structure from neuron firing \cite{boundy-singer,manfredi}, an idea that was inspired by the timbre-invariance observed in auditory perception \cite{saal2}. Utilising the speed-invariance of harmonic structure, here we considered artificial decoding models trained with augmented datasets where the held-out speed was simulated by stretching or compressing training data in the frequency domain. This augmentation method provided both a novel data generation procedure for training predictive models of texture class and resulted in an effective generalisation of those models across the speed of interaction.

Features derived from the frequency spectra of contact-induced vibrations have been used in robotics for texture classification experiments \cite{yi2017,sinapov2011,fishel}. However, our approach of decoding information from temporal artificial tactile data by using convolution neural networks with filters applied in the frequency domain that extract features from the shape of the frequency spectra, appears to be novel. Furthermore, speed-invariance of texture perception has been examined in robotics in only one other study \cite{romano2}, where stimulation speed was applied as an additional feature to a support vector machine classifier. By utilising a method based on a theory of speed-invariant texture perception in nature, our approach does not require any specific speed information. \textcolor{black}{One should also note that there are important differences from \cite{romano2} in the dynamics of the contact, in that they used a rigid tool running over a hard texture, whereas we used a soft fingertip interacting with a hard texture.}

\subsection{\textcolor{black}{Implications for robot touch}}

\textcolor{black}{Using natural touch as inspiration, in this paper we have identified several principles that are beneficial for artificial texture perception, arising from a consideration of the tactile cues from dynamic stimulation, transduction methods and encoding schemes. These techniques include the use of frictional cues for dynamic texture perception, the extraction of artificial RA-I tactile features from image differences, harmonic structure in a vibrotactile channel and speed-invariance as a data augmentation procedure. These methods will likely extend to other tactile sensors that provide rich spatio-temporal data, have a vibrotactile channel and can measure shear.}

\textcolor{black}{The present work opens up several areas for future exploration. The artificial \mbox{SA-I}, RA-I and vibrotactile/RA-II channels were considered in isolation, but how would they combine to give effective texture recognition? It can be straightforward to combine feature sets within deep neural networks, but there are subtleties in how multi-modal representations are learned. The stimuli used here varied in a single perceptual dimension, roughness, but how would the results extend to natural textures? The use of artificial stimuli from human psychophysics studies of other perceptual dimensions is a way forward to maintain the relation with human texture perception; however, methods not directly connected with human touch may be valuable too, depending on the application. Lastly, a multi-modal sense of touch involving slow-adapting, rapid-adapting and vibrotactile channels is valuable not just for texture perception, but also for other applications of artificial tactile sensing. Dexterous manipulation is one key application area, as it known from human studies that the Pacinian RA-II system is an integral component of how we pick up, handle and manipulate objects.}

\subsection*{Data Accessibilty}
All code to process the data, test/train the deep learning models and generate result figures is available at https://github.com/nlepora/afferents-tactile-textures-jrsi2022. Data are available at the University of Bristol data repository, data.bris, at https://doi.org/10.5523/bris.3ex175ojw0ckt25icp6g6p9j12.

\subsection*{Author Contributions}
N.P. and N.L. planned the research. N.P. conducted the experiments and analysis. N.P. and N.L. wrote the paper. N.L. did the revisions and final paper preparation.

\subsection*{Competing Interests}
We declare we have no competing interests.

\subsection*{Acknowledgements}
We thank the anonymous reviewers for their helpful and constructive advice. We also thank Stephen Redmond and Chris Kent for examining the PhD viva of N.P. upon which this paper is based. 

\subsection*{Funding}
This research was funded by a Leverhulme Research Leadership Award on `A biomimetic forebrain for robot touch' (RL-2016-39) and an EPSRC DTP PhD Scholarship for N.P.

\bibliographystyle{unsrt}
\bibliography{mybib}

\end{document}